\documentclass[letterpaper, 10 pt, conference]{ICRA/ieeeconf}  % Comment this line out if you need a4paper
\let\labelindent\relax

\IEEEoverridecommandlockouts                              % This command is only needed if 
\title{\LARGE \bf
SwapBot: Tool-Centric Robot Manipulation via Interchangeable End-Effectors and Tactile Sensing
}

\author{%
\makebox[\textwidth][c]{%
\shortstack[c]{%
{\normalsize\bfseries
Yunchu Zhang, Ray Song, Junhan Jiao,
Liyiming Ke, Yuchen Yang
}\\[1pt]
{\normalsize\bfseries
Ruihan Gao, Emma Romig,
Siddhartha Srinivasa, Yiyue Luo
}\\[5pt]
{\small\normalfont University of Washington}
}%
}%
}
\usepackage[dvipsnames]{xcolor} 
\usepackage{amssymb}
\usepackage{graphicx}
\usepackage{amsmath,amssymb,amsfonts}
\usepackage{enumitem}
\usepackage{algorithmic}
\usepackage{svg}
\usepackage{subcaption}
\usepackage{tcolorbox}
\usepackage{textcomp}
\usepackage{xcolor}
\usepackage{booktabs}
\usepackage{wrapfig}
\usepackage{caption}
\usepackage{hyperref} 
\usepackage{capt-of}       % for \captionof{figure}
\usepackage{cuted}         % provides the strip environment
\usepackage{booktabs,makecell,tabularx}

\renewcommand{\arraystretch}{1.2}
\newcommand{\best}[1]{\textbf{#1}}

\newcommand{\Method}{SwapBot}

\begin{document}

\maketitle
\thispagestyle{empty}
\pagestyle{empty}

%%%%%%%%%%%%%%%%%%%%%%%%%%%%%%%%%%%%%%%%%%%%%%%%%%%%%%%%%%%%%%%%%%%%%%%%%%%%%%%%
% \begin{figure*}[!t]
%     \centering
%     \vspace{-1em}
%     \includegraphics[width= \linewidth]{figures/Overview_New.pdf}
%     \caption{Figure one for wow tasks with teleop and policy}
%     \label{fig:Overview}
% \end{figure*}
% \twocolumn[{%
% \renewcommand\twocolumn[1][]{#1}%
% \maketitle
% \vspace{-0.5in}
% \begin{center}
%     \centering
%     \captionsetup{type=figure}
%    \includegraphics[width=1.0\textwidth]{figures/teaser_v3_compressed.pdf}
%      \vspace{-0.2in}
%     \caption{We deploy \ourmethod on a Unitree G1 humanoid to demonstrate diverse whole-body loco-manipulation capabilities. (\textbf{a}) carrying 4.5kg heavy payloads via whole-body contact; (\textbf{b}, \textbf{c}) expressive motion while carrying box; (\textbf{d}) general humanoid object interaction beyond manipulation; (\textbf{e}, \textbf{f}) carrying irregularly shaped heavy objects with instance-level generalization.
% }
%     \label{fig/teaser}
% \end{center}
% \vspace{-0.05in}
% }]
\begin{strip} 
\centering 
\vspace{-3.0\baselineskip} % (optional) tighten space below 

\includegraphics[width=0.95\textwidth]{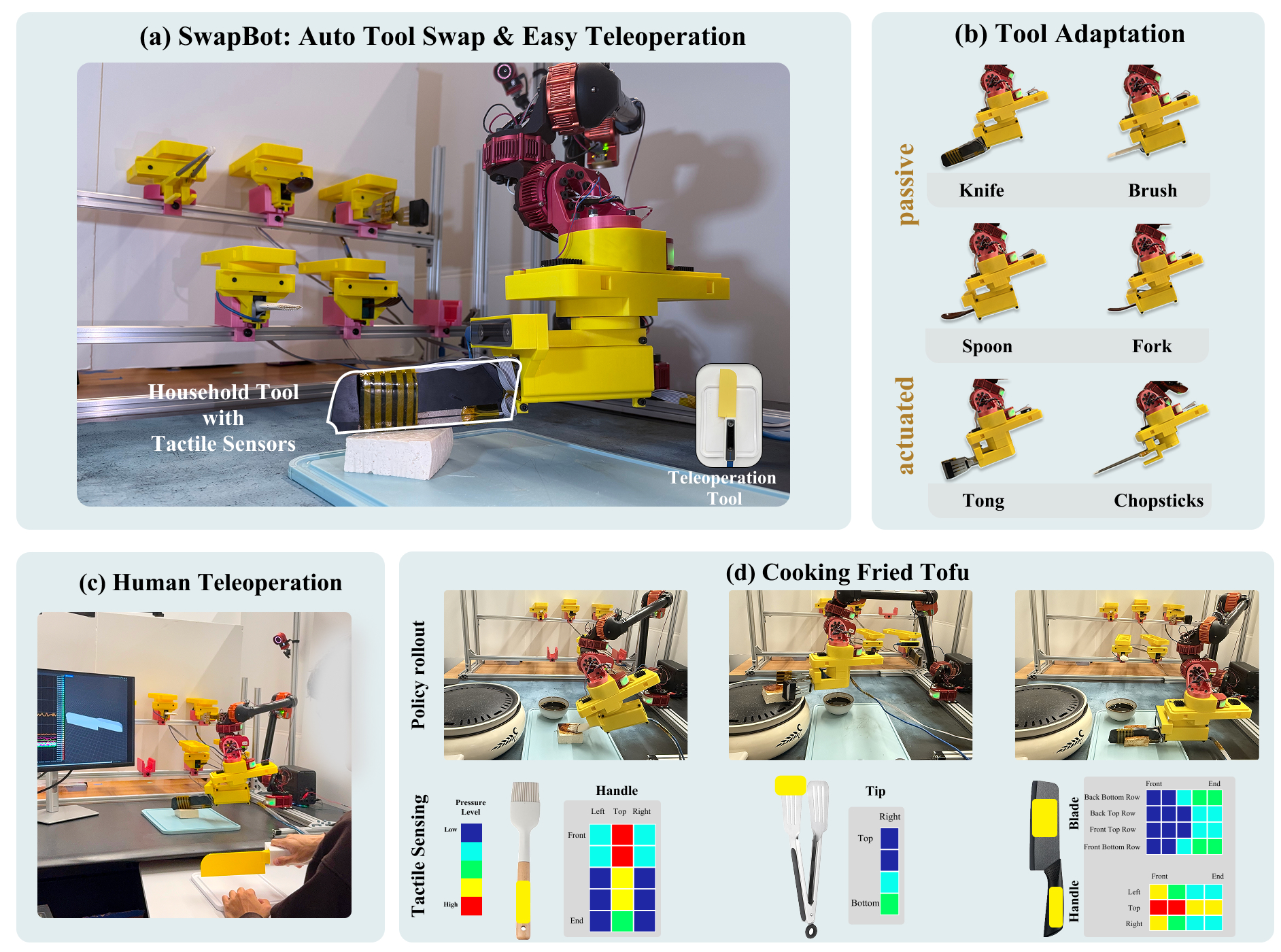} \captionof{figure}{SwapBot is a modular, tool-centric end effector system. (\textbf{a}) It supports autonomous tool swapping, integrates tactile sensors, and enables intuitive teleoperation.  (\textbf{b}) Low-cost 3D-printed tool adapters convert everyday household tools to robot end effectors. (\textbf{c}) Human teleoperation with visualized tactile feedback enables natural control and efficient demonstration collection. (\textbf{d}) \Method\ autonomously executes a long-horizon cooking task: brushing marinade with a brush, grasping tofu with tongs, and slicing with a knife, with tactile sensing (visualized as color heatmaps) and RGB inputs. 
% This demonstrates seamless tool-swapping and multi-step policy execution.
} \label{fig:Overview}

\vspace{-1\baselineskip} % (optional) tighten space below 
\end{strip}
\begin{abstract}

Robot manipulation in everyday environments inherently involves substantial tool usage. Seemingly simple household tasks, such as cooking pan-fried tofu, require forceful cutting, delicate brushing, and precise placement using different tools under contact-rich interaction. Thus, the ability to autonomously swap and use task-appropriate tools is important for tool-centric manipulation. We introduce \Method, a modular, tool-centric end effector system featuring low-cost adapters that quickly convert common household tools into robot end effectors. Rather than pursuing a single universal gripper, \Method\ adopts a unified design principle that combines a fixed robot-side coupling interface with low-cost, tool-side adapters, enabling compatibility with diverse tools—including articulated ones—and multiple robot platforms. Each adapter integrates compact tactile sensors, supports fast autonomous swapping, and interfaces with an intuitive teleoperation system for efficient demonstration collection. User studies show that, during teleoperation, \Method\ reduces task completion time and operator workload. Using collected demonstrations on the robot, we train imitation learning policies to successfully perform tool-centric tasks like cooking tofu, stirring salad, flipping meat. Our ablation study shows that using tactile-augmented demonstrations improves policy performance across contact-rich skills, increasing success rates by 15--25 percentage points over RGB-only policies. These results highlight modular, tactile-equipped tools as a practical and extensible pathway toward robust and capable robot manipulation. Project website \href{https://personalrobotics.github.io/SwapBot/}{https://personalrobotics.github.io/SwapBot}.
% solve tasks like feeding/repairing/ cooking/fine assembly
% what's big challenge
% 1)dextrous
% 2)multiple-tools
% 3)forceful contact
% 4)learning from human is useful

% cook/surgical/ based on those requirement-->leads to this current design, swapable/measure/capable of dextrous

% why intersting: low cost/modular/open
% actuate without adding motor to each things

% why this sensing:
% use joint/torque measure
% 6axis measurement

% open/low cost/extensible
% something did not work and how to make it work

% how to select the material/sensibility lenght
% conductive facbirc vs conductive thread
% why this is better compared to ...

% Design choice section:
% swap and sensing
% journey

% system analysis

\end{abstract}

%%%%%%%%%%%%%%%%%%%%%%%%%%%%%%%%%%%%%%%%%%%%%%%%%%%%%%%%%%%%%%%%%%%%%%%%%%%%%%%%
\vspace{-0.5em}
\section{Introduction}
\begin{figure*}[!ht]
    \centering
    \includegraphics[width=0.85\textwidth]{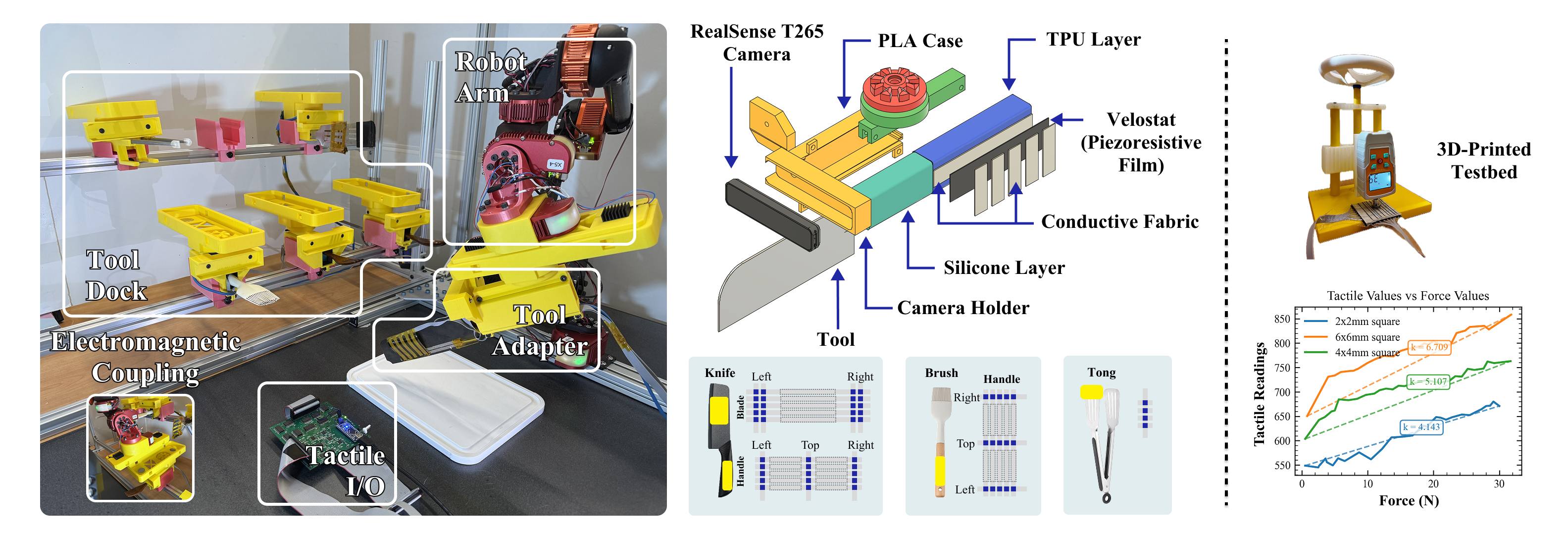}
    \caption{
    (\textbf{Left}) SwapBot setup overview with the robot arm, robot-mounted tool adapter, electromagnetic coupling, the tool dock, and tactile I/O board. 
(\textbf{Middle}) The structure of the instrumented tools with integrated tactile sensing using piezoresistive film, showing sensor layers and placements, with examples for a knife, brush, and tongs. (\textbf{Right}) Characterization of tactile sensors, showing force–response curves for electrodes of different sizes. Larger electrode areas exhibit higher sensitivity, but also cause non-smooth signals with sudden jumps under load. Smaller electrodes have smoother signals and a wider measurable range.}
    \label{fig:System}
\vspace{-1.8em}
\end{figure*}
% intuitive data collection
Robotic manipulation is advancing toward increasingly human-level capabilities; however, seemingly simple tasks, such as cooking pan-fried tofu, remain challenging for robots to perform. Performed with different tools under contact-rich interaction, these tasks require precise but forceful control (cutting), delicate interaction with deformable objects (brushing), and dexterous motion (grasping and placing). Such \textit{tool-centric manipulation} is challenging under the current end-effector design, which falls into two broad categories. First, \textit{anthropomorphic hands} mimic human dexterity~\cite{shaw2023leap, shadowhand}, but their mechanical complexity imposes a high cost in maintenance, sensing, and control, hindering scalable data collection and policy learning. Second, \textit{parallel jaw grippers}, though practical and demonstrating impressive versatility~\cite{zhao2024aloha, black2024pi_0, khazatsky2024droid}, struggle to perform tasks requiring nuanced physical interactions, such as cutting or spreading. These capability vs controllability tradeoffs create significant barriers to more capable and scalable tool-centric robot manipulation.

Humans accomplish complex manipulation tasks by using tools, e.g., knives, brushes, or screwdrivers, as extensions of their bodies. Robotists have explored tool use using dexterous hands ~\cite{chen2025vegetable, yin2025dexteritygen} or parallel grippers~\cite{lin2022planning, shi2023robocook}. However, existing methods either inherit the complexity of manipulating anthropomorphic hands, struggling to maintain tool stability under load, or, critically, lack tactile sensing~\cite{zhao2023learning} at the tool-object interface, hindering effective control and learning for contact-rich tasks.

%Tool use is also a hallmark of intelligence, distinguishing humans from animals and enabling robots to extend their abilities for complex manipulation. 
%Our work focuses directly on the critical interaction and manipulation parts of the task. 
To balance robot manipulators' capability and controllability, we propose to equip robots with the ability to swap modular tools through a tool changer. Industrial tool changers, such as those used in CNC machining or on UR5 robots, demonstrate the practicality of extending robotic capabilities through automatic tool exchange. However, these costly systems are typically customized for specific robots and cannot easily accommodate actively articulated tools - e.g., tongs and scissors. Moreover, they lack integration with tactile sensing, limiting their ability to support contact-rich manipulation tasks. Instead, our approach turns accessible household items into robot tools and equips each tool with compact, custom tactile sensors, enabling robots to both select the appropriate tool for each subtask and sense contact at the tool–object interface. Drawing inspiration from surgical robots that interchange end-effectors such as scissors, cauterizers, and forceps, \textit{our system thus enables scalable household manipulation through tool diversity, not anatomical mimicry}. 

We present \textbf{\Method, a modular, low-cost, tool-centric manipulation system with integrated tactile sensors}, as shown in Fig.~\ref{fig:Overview}. It features: (1)a unified adaptation interface for diverse tools and robots,  (2) fast and autonomous tool swapping protocol, (3) integrated tactile sensing,  and (4) intuitive teleoperation for demonstration data collection. Our system employs inexpensive, 3D-printed interfaces (\$10 each) \textit{to quickly adapt common household tools}. We apply this modular system to three robot platforms: Franka~\cite{franka2019}, UR5~\cite{ur5}, and Tycho~\cite{ke2021grasping} and to a variety of household tools - including articulated tools that require actuation, such as tongs and chopsticks. Robots equipped with our system can use our designed coupling interface and protocol \textit{to autonomously switch the tool within 10 seconds}. Our teleoperation interface combines tool-matched leader devices with real-time visualized tactile feedback, providing operators with an intuitive interface for contact-rich manipulation. Furthermore, we show that employing tactile sensing significantly boosts the performance of the autonomous policy learned from human demonstrations.

\begin{figure}[!t]
    % \vspace{-1em}
    \centering
    \includegraphics[width=\columnwidth]{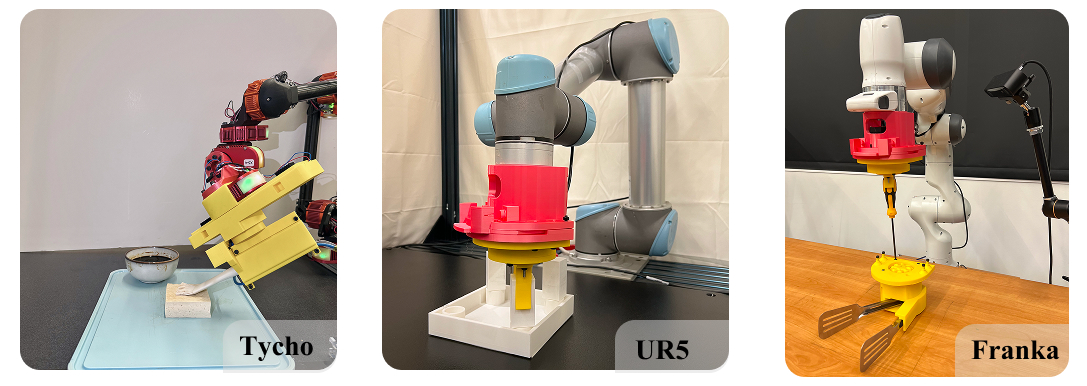}
    \caption{SwapBot can be adapted to three different types of robots: Tycho, UR5, and Franka. 
    % The modular, tool-centric manipulation system with integrated tactile sensing enables complex manipulation.
    }
    \label{fig:multi-robot}
    \vspace{-2.5em}
\end{figure}

To assess the ease of teleoperation with \Method, we conduct human studies with 12 participants of varying teleoperation skill levels (beginner, intermediate, and expert), focusing on two representative tasks: brushing and cutting. Participants report less subjective workload of control and higher success rates compared to the use of a generic teleoperation tool. \Method\ also enables the teleoperated completion of tasks previously found challenging or impossible for simple grippers, reducing task completion time and offering more satisfying user experiences. In robotic experiments, we further demonstrate the autonomous execution of a challenging long-horizon task: cooking pan-fried tofu. Using the demonstration data that our system collected, we train imitation learning policies and show that tactile feedback significantly improves policy performance, increasing task success rates by 25 absolute percentage points, from $35\%$ to $60\%$. We additionally provide qualitative results (e.g., stirring salad and flipping meat) in the supplementary videos. In summary, our contributions include:
\begin{enumerate}[noitemsep, topsep=0pt, leftmargin=*]
    \item A \textbf{unified, low-cost, modular tool-centric end-effector system} that supports rapid autonomous tool swapping and the use of diverse household tools, including articulated tools.
    \item \textbf{Integrated compact tactile sensing} that improves both teleoperation performance and imitation-learning success in contact-rich task.
    \item An \textbf{intuitive teleoperation interface for collecting demonstrations}, improving operator precision and enabling high-quality data collection.
    \item \textbf{Empirical evaluation} via user studies and policy learning: quantitative tofu-cooking results and qualitative demonstrations of stirring salad and flipping meat, showing improved user experience and policy performance.
\end{enumerate}

\section{The SwapBot System}
\begin{figure*}[!ht]
    \centering
    \includegraphics[width=0.95\textwidth]{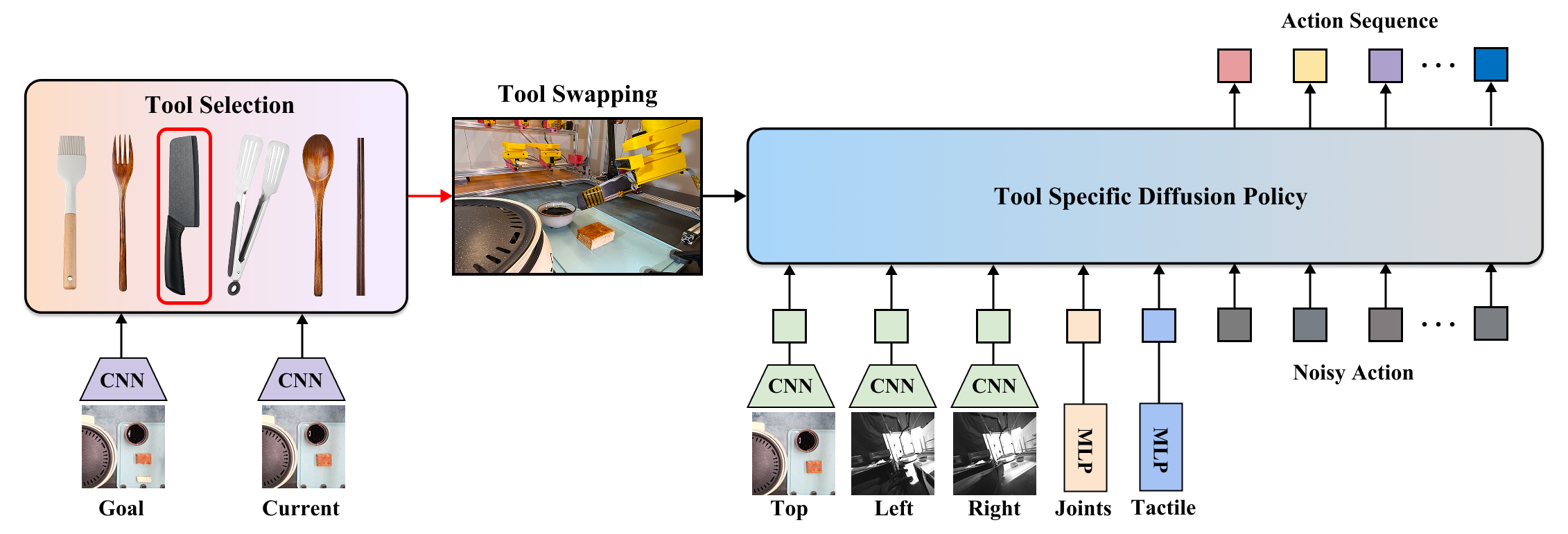}
    \caption{
    Overview of the hierarchical policy learning framework. 
    The high-level tool-selection policy $\pi_{\text{sel}}$ takes top-view observations $\mathcal{O}_{top}$ and a goal image $\mathcal{O}_g$ as input and outputs a distribution over candidate tools. The selected tool index $\tau$ activates the corresponding low-level tool-specific control policy $\pi_{\text{tool}}^\tau$. Each policy conditions on visual observations $\mathcal{O}_t$, tactile feedback $\mathcal{T}_t$, and proprioceptive state $s_t$ to generate end-effector actions. During long-horizon execution, the robot continuously conducts tool selection and skill execution in a closed-loop, which enables \Method\ to compose tool-specific skills for long-horizon tasks.
    }
    \label{fig:method}
    \vspace{-2em}
\end{figure*}
The \Method\ system contains four main components: modular and interchangeable end-effectors (Section \ref{sec:2a}), compact tactile sensing (Section \ref{sec:2b}), the teleoperation interface (Section \ref{sec:2c}), and policy learning from demonstrations (Section \ref{sec:2d}). Our plug-and-play design supports low-cost prototyping with a variety of daily-use tools while supporting tactile sensing, large-scale demonstration collection, and downstream policy learning. We describe each component below. 

\subsection{Modular End-Effectors:
Tool Adapters and Coupling} \label{sec:2a}

\textbf{Design Goals.} Industrial robotic coupling is usually tailored to specific tools and robots and often costs hundreds or even thousands of dollars. In contrast, we aim to develop a modular tool coupling system that is accessible and low cost (tool-side modules under \$10), rapid swapping of tools, robot-agnostic, enabling compatibility with common manipulators, and flexible for both passive and articulated tools. To achieve these goals, we design our tool-swapping adapter and coupling system to enable a fully accessible fabrication workflow and straightforward assembly using basic hand tools, such as an Allen wrench. 

%sandy{Ensure the following is the correct figure reference given upstream text changes.}
As shown in Figure \ref{fig:System} (left), we 3D-print tool adapters and inserts to enable the quick instrumentation and mounting of common household tools. The adapter both attaches the tool securely and houses the tactile sensors stably. For tools with moderate handle sizes, the same adapter can be reused, while soft silicone fills the remaining gap to accommodate variations in handle shape and size. Only substantially thinner tools require a smaller 3D-printed adapter, while the robot-side coupling interface remains unchanged. Though many everyday tools include their own actuation mechanisms, equipping each with dedicated motors for automatic control would be complex and expensive. Instead, we shift all power and control functionalities to the robot side, ensuring a simpler and lower-cost tool implementation. To transmit torque, we add a bearing system on top of the adapter cover that transfers motion from the robot’s end joint. For robots without a final  joint, we supplement with a single servo motor to ensure compatibility across Franka, UR, and other common platforms (Fig. \ref{fig:multi-robot}). 

To support rapid tool exchange, we implement a self-aligning electromagnetic coupling that mates a robot-side hub with tool-side adapters. This design supports both passive tools (e.g., knives, brushes) and articulated tools (e.g., tongs, chopsticks) through a pass-through mechanism. A kinematically guided gear ensures reliable engagement and release under small alignment errors, while the electromagnet provides sufficient holding force for secure tool attachment during forceful manipulation. In practice, our system completes a swap in about 10~s by guiding the robot to a docking station, releasing the current tool, and attaching a new one. This approach shifts capability from a single complex hand to a library of instrumented tools while retaining the ability to compose skills across tools. We summarize key design specifications in Tab.~\ref{tab:specs}.

\begin{table}[t]
\centering
\caption{SwapBot End-Effector Specifications}
\label{tab:specs}
\begin{tabular}{p{3.2cm} p{4.5cm}}
\toprule
\textbf{Feature} & \textbf{Specification} \\
\midrule
Fabrication Accessibility & 3D-printed PLA, TPU, and Silicone, assembled with basic hand tools \\
Compatibility & Franka, UR, Tycho and other common platforms \\
Tool Support & Passive (knives, brushes, forks, spoons) and articulated (tongs, chopsticks) \\
Coupling \& Tool Swapping & Self-aligning electromagnetic hub with kinematic gear \\
Retentive Force & Up to 50~N via electromagnet \\
Swap Time & $\sim$10~s with docking procedure \\
Tool-side Cost & $\sim$\$3–4 (3D printing filaments), \$2–3 (screws/standoffs), $\sim$\$2 (sensors materials)
\\Shared robot-side Cost& \$20 (electromagnet system) \\
\bottomrule
\end{tabular}
\vspace{-2em}
\end{table}

\subsection{Compact Tactile Sensing: Instrumented Tools}
\label{sec:2b}
We design compact, tool-specific piezoresistive tactile sensors that provide task-relevant contact information for teleoperation and policy learning. By tailoring the sensor layout to each tool, we capture force information without adding bulk or compromising usability. 

The piezoresistive tactile sensors consist of a three-layer structure, shown in Fig.~\ref{fig:System} (center), where a force-sensitive material layer (Velostat) is sandwiched between top and bottom electrodes (conductive fabric) for signal transmission. These sensors ~\cite{sundaram2019learning} convert real-time normal force pressure information into an electrical signal via the drop of resistance. The electrodes are made from copper-nickel conductive fabric, laser-cut to precise dimensions for optimal fit considering the sensitivity they need for tasks. To accommodate the limited surface area of tools, we implement the sensors in a matrix layout, where sensing nodes are formed at the intersections of orthogonal electrode rows and columns. This configuration enables an n $\times$ n sensor array using only 2n electrodes. Our approach is low-cost, relies on only off-the-shelf materials, and supports easy customization through digital fabrication.

We show the design and placement of tactile sensors for each tool in Fig.~\ref{fig:System} (center). For tools with larger surfaces, we place sensors on both the handle and the blade or tip (e.g., tongs, spoons, knives). For smaller or thinner tools, such as chopsticks, brushes, and forks, it is impractical to mount sensors directly on the blade or tip. In these cases, we instrument only the handle, providing indirect measurements of tool–object interaction. This strategy ensures that tactile information is captured where feasible while keeping the modules compact and lightweight. For blade or tip instrumentation, we mount the fabricated tactile sensors directly onto the tool surface. To accommodate handles of varying sizes, instead of attaching sensors directly, we install them between the TPU layer and the PLA adapter cover and fill the gap between the handle and TPU shell with soft silicone. This approach allows the sensors to comply with different handle shapes, stabilizes the installation process, and ensures consistent reflection of sensor readings during interaction.
\begin{figure*}[!t]
    \centering
    \includegraphics[width=\textwidth]{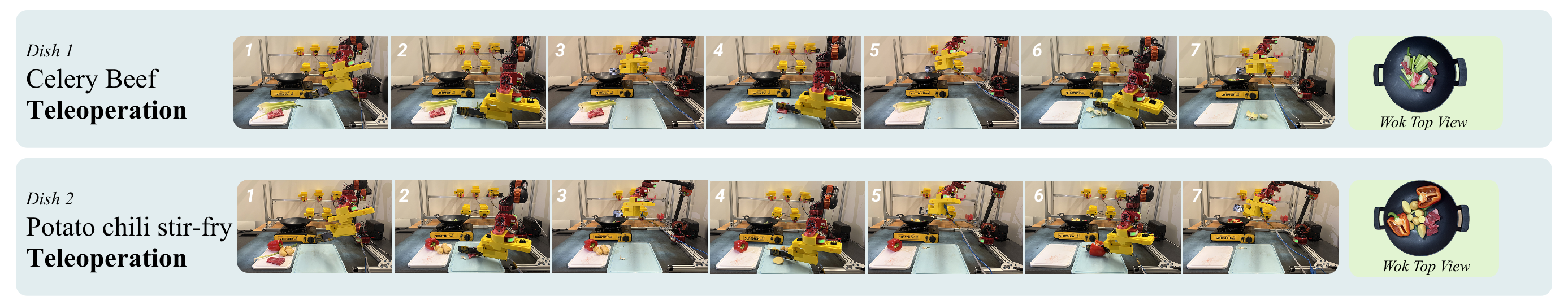}
    \caption{Tool-centric teleoperation for two challenging and complex long-horizon cooking tasks: Dish~1: Celery Beef and Dish~2: Potato and Pepper Stir-Fry.}
    \label{fig:Tasks}
    \vspace{-2em}
\end{figure*}
\subsection{Teleoperation Interface: High-Quality Demonstration Collection}
\label{sec:2c}

We design the teleoperation interface to maximize operator intuitiveness and data quality, both of which are critical for downstream policy learning. Our interface maps the intuitive leader device's 6-DoF pose and actuation commands directly to the corresponding tool on the robot side. By using similar tools on both sides, e.g., chopsticks to control chopsticks or a knife to control a knife,  operators interact with familiar household tools, making control feel natural. This reduces the re-targeting effort common in teleoperation and lets operators focus on task execution rather than compensating for mismatched dynamics. For each leader tool, we estimate the 6-DoF pose using an Intel RealSense T265 tracking camera, and use the result to drive the robot during teleoperation. For articulated tools, we infer the actuation command from the relative motion measured by two IMUs, providing accurate correspondence between human and robot tool motion.

During demonstration data collection, all data streams are synchronized at 20~Hz, including multi-view RGB images, tactile signals, robot joint states, actions, and task labels. This ensures that multimodal inputs remain temporally aligned for learning. We also provide visual overlays that display the live contact forces and tactile maps to help operators anticipate and refine contact-rich interactions. As we will show in Fig.~\ref{fig:UserStudy}, tactile information is visualized on-screen to indicate contact information. This immediate feedback provides operators with additional information about contact, improving both control precision and the quality of collected demonstrations. 

To evaluate the interface, we conduct controlled human studies to measure task completion time, success rate, and the subjective workload (see Sec.~\ref{exp:userstudy}). Integrated tactile feedback emphasizes subtle events, such as first contact or changes in load distribution, which are often ambiguous in vision alone. This design reduces cognitive load and improves operator precision for long-horizon tasks such as cutting, grasping, and transporting.

\subsection{Policy Learning: From Demonstrations to Long-Horizon Manipulation} \label{sec:2d}
Inspired by RoboCook~\cite{shi2023robocook} and shown in Fig.~\ref{fig:method}, we train hierarchical policies from demonstrations that scale to long-horizon tasks. The policies are divided into two components: (1) the \textit{high-level tool-selection policy $\pi_{\text{sel}}$} determines which tool to use at each stage of the task, and (2) the \textit{low-level, tool-specific control policies $\pi_{\text{tool}}^i$} are trained to execute fine-grained manipulation skills conditioned on visuo–tactile inputs. We assume that long-horizon tasks can be segmented into subtasks with a subgoal available for each stage. In our current implementation, the sequence of subgoal images is manually specified before execution. We formulate the problem as follows: given top-view RGB observation $\mathcal{O}_{top}$ and a goal image $\mathcal{O}_{g}$, the tool-selection policy $\pi_{\text{sel}}(\mathcal{O}_{top}, \mathcal{O}_{g})$ predicts a probability distribution over candidate tools. We implement $\pi_{\text{sel}}$ as a classifier that uses a ResNet feature extractor followed by an MLP with a SoftMax output layer. The selected tool index $\tau$ specifies which low-level policy to activate. The corresponding control policy $\pi_{\text{tool}}^\tau$ then takes as input the current multi-view images $\mathcal{O}_{t}$, tactile signal $\mathcal{T}_{t}$, and robot proprioceptive state $s_{t}$ and generates robot actions $a_t = \pi_{\text{tool}}^\tau(\mathcal{O}_{t}, \mathcal{T}_{t}, s_{t})$. We implement $\pi_{\text{tool}}$ using a U-net based diffusion policy~\cite{chi2025diffusion} that generates smooth sequences of end-effector commands. 

During long-horizon task execution, the active policy is considered to have completed its skill when the Euclidean distance between consecutive end-effector positions remains below 0.005 m for 20 consecutive control steps. The robot then resets to its initial pose to avoid occluding camera views. The selector $\pi_{\text{sel}}$ then takes the current observation with the new goal image to choose the following tool, ensuring continuous task execution through visual feedback and hierarchical decision making. This framework enables \Method\ to successfully execute long-horizon tasks, such as cooking pan-fried tofu, by sequencing tool-specific policies into a coherent task plan.

\section{Experiments}
% \vspace{-0.70em}
\begin{figure*}[!t]
    \centering
    \includegraphics[width=0.95\textwidth]{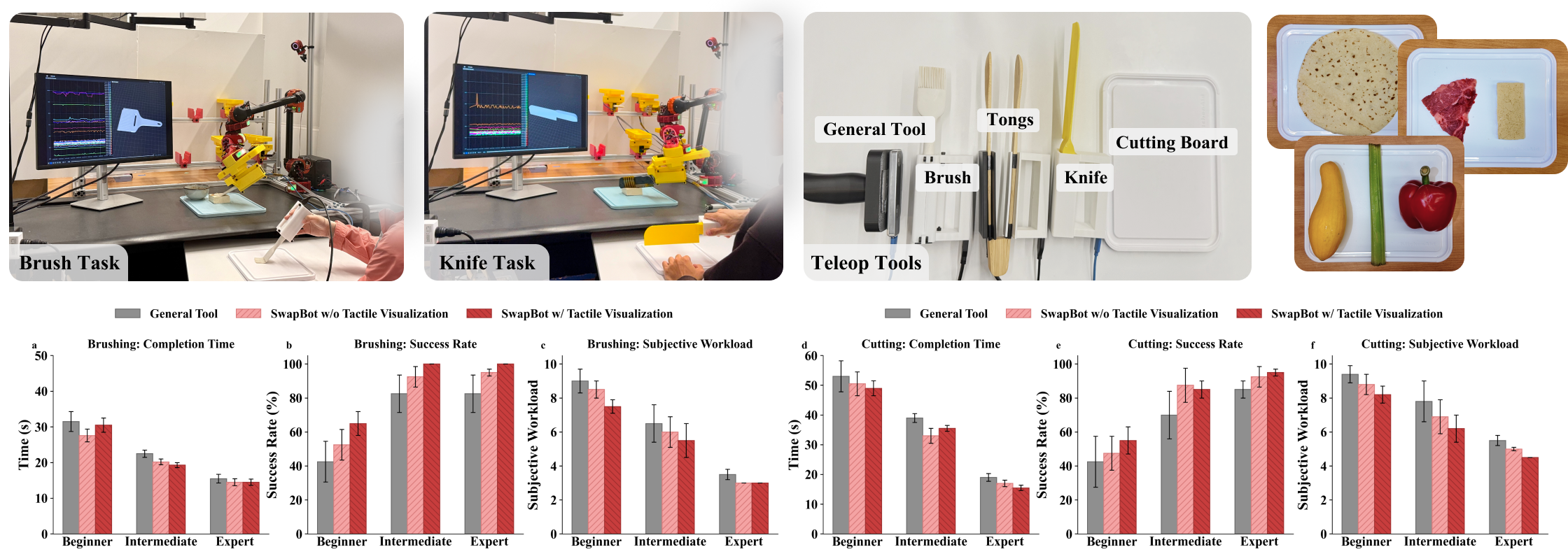}
    \caption{User study results for brushing (a--c) and cutting (d--f) tasks across three skill levels of participants. Error bars indicate 95$\%$ confidence intervals across participants. We compare a general teleoperation tool, SwapBot without visual tactile feedback, and SwapBot with real-time visualized tactile feedback. Overall, the SwapBot interface yields faster completion times, higher success rates, and lower subjective workload than the general teleoperation tool baseline.}
    \label{fig:UserStudy}
    % \vspace{-1.2em}
\end{figure*}

Our experiments aim to answer the following questions. \textbf{(1) Characterization of tactile sensing:} 
How well do our compact, tool-specific tactile sensors perform in terms of sensitivity and detection range, and does our design achieve the right balance for integration into diverse tools? \textbf{(2) Intuitive teleoperation:} 
Does our tool-centric teleoperation interface provide a natural, user-friendly experience that improves demonstration efficiency and data quality compared to standard baselines? 
\textbf{(3) Policy learning with tactile feedback:} 
Does integrating tactile sensing into imitation learning policies improve performance on complex, long-horizon tasks?

\subsection{Characterizing Tool-Specific Tactile Sensing}
Directly applying the tactile sensors without considering their size and sensitivity can lead to saturated and inconsistent signals during data streaming. To ensure smooth and reproducible sensing, we characterize the performance of our resistive tactile sensors made with electrodes of varying widths. Signals were recorded under controlled pressure using a force gauge, from which we extracted both sensitivity and detection range. These measurements provide valuable insights into how the electrode design impacts sensing quality and guides our choice of parameters.

As we show in Figure \ref{fig:System} (right), electrode size has a direct impact on both sensitivity and signal quality. Larger electrodes provide higher sensitivity, with identical forces producing stronger responses due to increased overlap between the conductive layers. However, these larger electrodes also generate non-smooth signals with sudden jumps under load, and can saturate the 10-bit analog-to-digital converter (ADC), making them less reliable for continuous feedback. Smaller electrodes, by contrast, sacrifice sensitivity but produce smoother, more stable signals with a wider measurable range. To balance sensitivity, detection range, and signal fidelity within ADC limits, we select a 2 mm electrode width, which provides sufficient sensitivity while maintaining smooth, stable responses.
\vspace{-0.5em}

\subsection{Evaluating Intuitive Tool-Centric Teleoperation}
\label{exp:userstudy}
To evaluate the interface, we conduct a controlled human study with 12 participants spanning three teleoperation skill levels (beginner, intermediate, and expert), with four participants in each group. Each participant performs both brushing and cutting tasks under three teleoperation conditions: a general VR-style tool, SwapBot without tactile feedback, and SwapBot with visualized tactile feedback. Each participant performs 10 trials per task–condition pair, terminating upon success or a 60-s timeout. Completion time is computed over successful trials, while failures contribute only to the success-rate metric. The order of teleoperation conditions was counterbalanced across participants to reduce learning and fatigue effects, while task order was randomized. We measured task completion time, success rate, and subjective workload using NASA-TLX, normalized to a 0–10 scale. For completion time and success rate, we first aggregate the 10 trials for each participant and then report the mean performance across the four participants within each expertise level.

As demonstrated in Fig.~\ref{fig:UserStudy}, the SwapBot interface generally improves teleoperation performance over the baseline across both tasks. The visualized tactile feedback provided operators with direct information about contact forces and load changes, enabling more effective manipulation during teleoperation. This interface also supports demonstration collection for more complex tasks, as illustrated in Fig.~\ref{fig:Tasks}.

Quantitatively, participants consistently achieved higher task success rates and reported lower cognitive load with our system compared to the baseline general teleoperation tool. These results suggest that our teleoperation design supports efficient and intuitive demonstration collection.
\begin{figure*}[!t]
    \centering
    % Subfigure a: Brushing
    \begin{subfigure}[t]{0.32\textwidth}
        \centering
        \includegraphics[width=\linewidth]{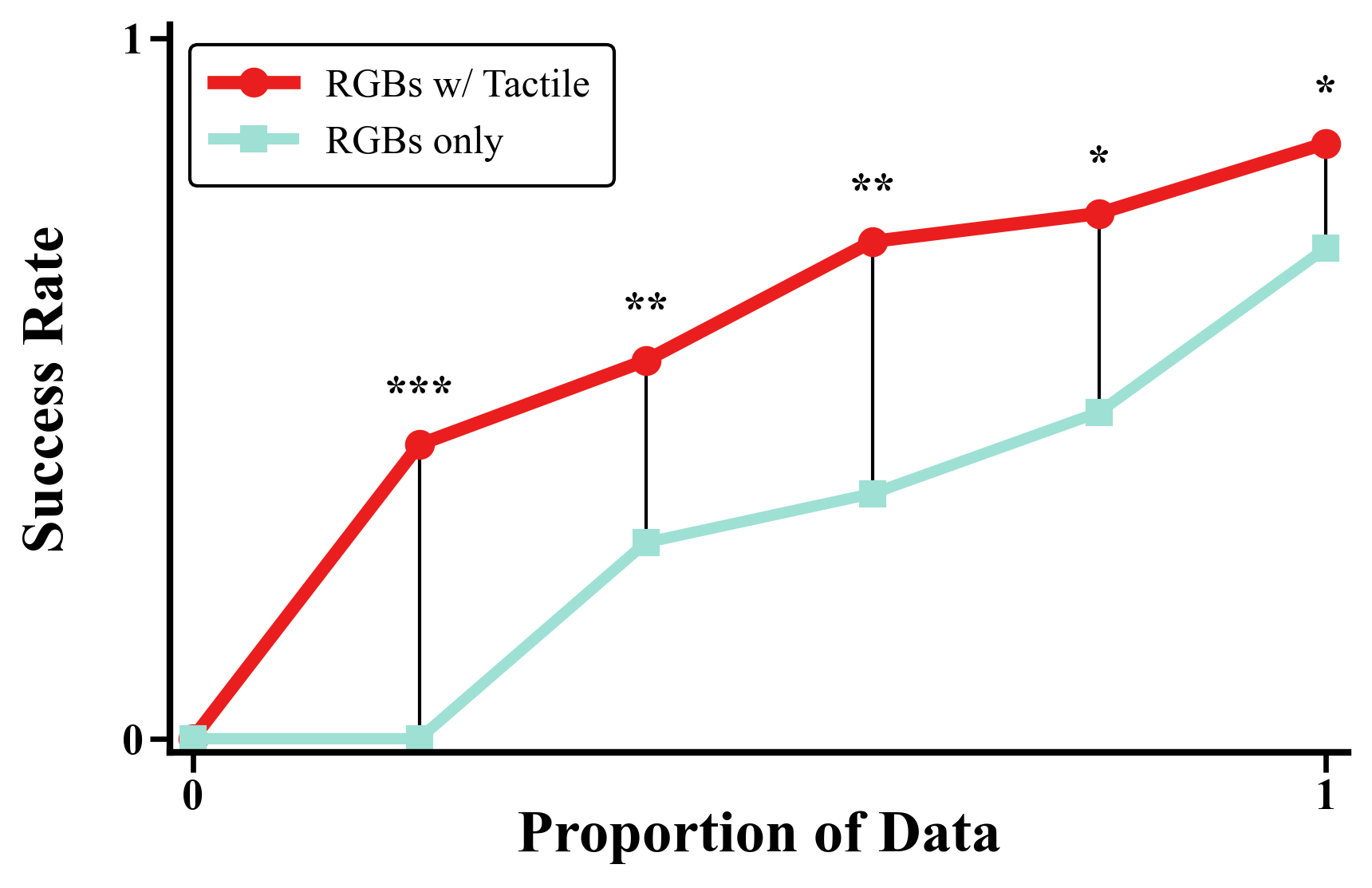}
        \caption{Brushing task}
        \label{fig:brush}
    \end{subfigure}
    \hfill
    % Subfigure b: Transporting
    \begin{subfigure}[t]{0.32\textwidth}
        \centering
        \includegraphics[width=\linewidth]{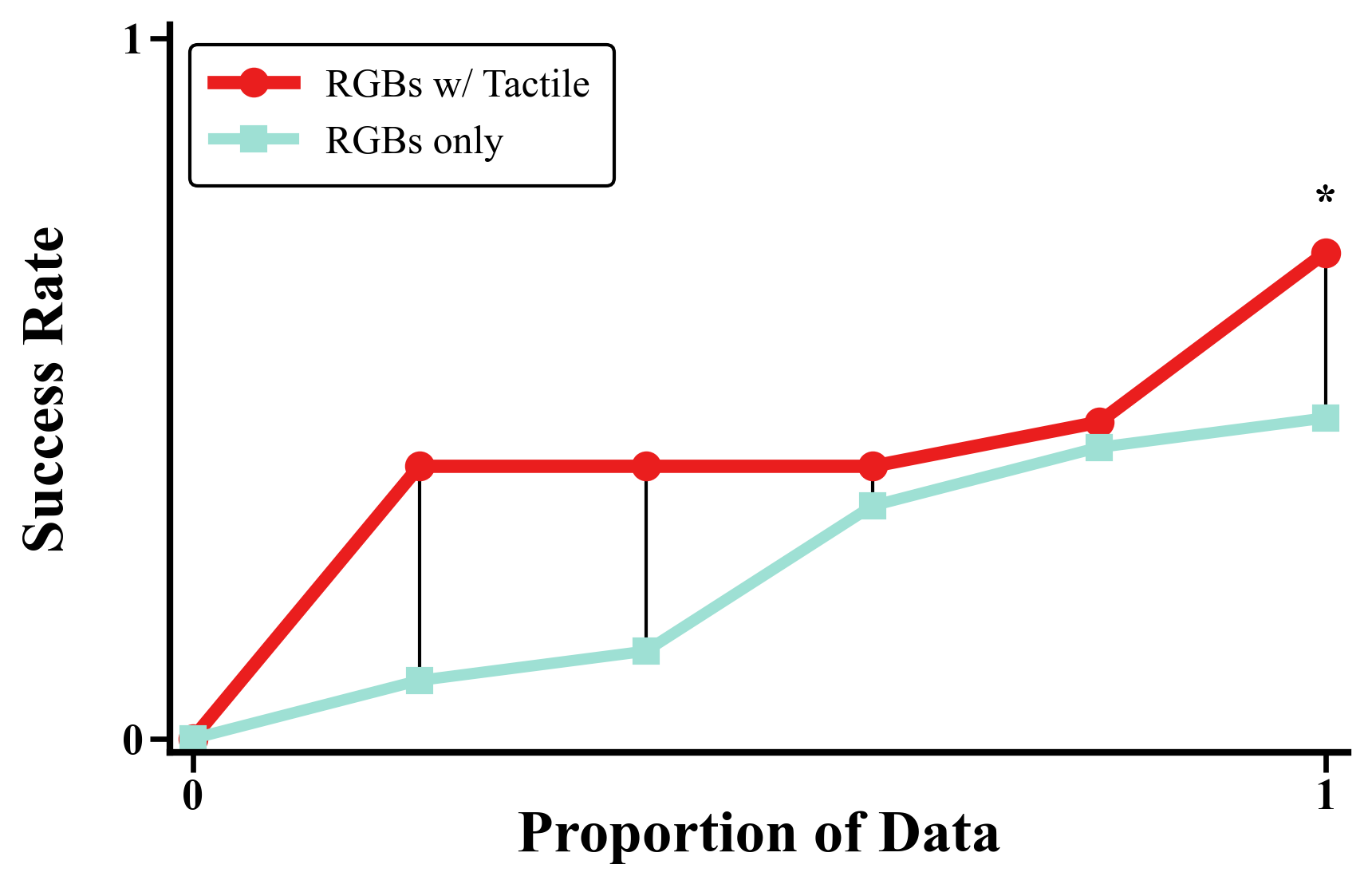}
        \caption{Tong picking task} 
        \label{fig:transport}
    \end{subfigure}
    \hfill
    % Subfigure c: Cutting
    \begin{subfigure}[t]{0.32\textwidth}
        \centering
        \includegraphics[width=\linewidth]{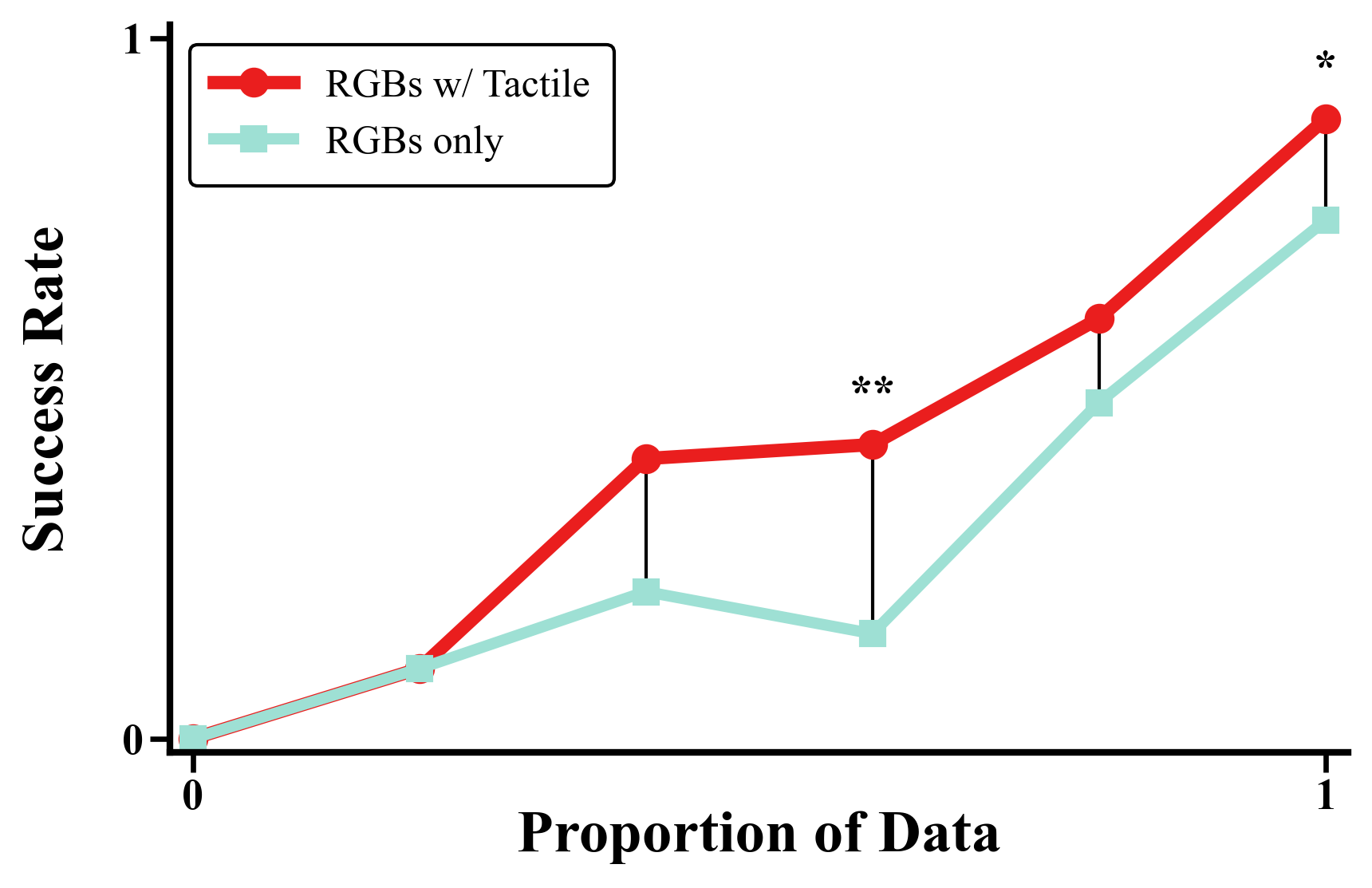}
        \caption{Knife cutting task}
        \label{fig:cut}
    \end{subfigure}
\caption{\textbf{Policy learning performance with and without tactile sensing.}  We average success rates over 20 evaluation rollouts per seed and run three random seeds; we then apply paired $t$-tests for significance ($*$ 
$p<0.1$, $**$ $p<0.05$, $***$ $p<0.01$). As the amount of data increases, tactile signals consistently provide a boost over RGB-only baselines and yield larger gains in low-data regimes.}
    \label{fig:t-test}
    % \vspace{-1.5em}
\end{figure*}
\vspace{-.5em}
\subsection{Policy Learning with Tactile Feedback for Long-Horizon Tasks}

We then evaluate whether tactile sensing improves imitation learning for complex, long-horizon manipulation. As a testbed, we use the challenging real-world task of preparing pan-fried tofu for quantitative experiments. The tofu is fragile and easily breaks if mishandled, requiring both gentle and dexterous manipulation. The robot must (1) gently brush sauce onto the tofu, (2) transport it to a pot using tongs for frying, and (3) return it to the cutting board and slice it into two pieces without damage. A trial is successful if the tofu is placed back on the board and cut cleanly into two pieces. Additionally, we provide qualitative demonstrations of other tool-mediated behaviors (e.g., stirring salad and flipping meat) in supplementary videos.

We used our teleoperation interface to collect 100 demonstrations for each stage, i.e., brushing, transporting to the pot, and cutting. The reverse transport back to the board was not evaluated quantitatively since it mirrors the forward motion, but we collected the same number of demonstrations to train the policy for completeness and composition. We then used these demonstrations to train tool-specific imitation learning policies under different input modalities. As baselines, we compared against policies trained without tactile sensing. We also performed ablations using: tactile only, top-down RGB + tactile, and egocentric T265 RGB + tactile. For all policies, we used a CNN feature extractor concatenated with proprioceptive state and tactile signals (when available) as inputs to the policy network. We evaluated each tool-specific policy across 20 trials in a 20~cm~$\times$~20~cm workspace. Policies trained with tactile sensing achieved consistently higher success rates and were robust to small tofu shape changes and variations on initial position. RGB-only policies, in contrast, frequently failed due to partial observability of the tool--object interaction---especially at the tong tips where contact onset and incipient slip are often occluded in RGB---leading to poor contact regulation, slipping, or damage. Egocentric RGB + tactile performed poorly, reaching at most 40\% success even with 100 demonstrations and staying near zero in tong picking and knife cutting tasks. Top-down RGB + tactile achieved moderate success (up to 50\% in brushing and cutting), but it lagged significantly behind our full RGBs + tactile method. 

Though the top-down view offers reliable localization of the tofu on the cutting board, it lacks local visual detail for fine control. Conversely, the egocentric T265 view provides local information but suffers from its angled viewpoint (greater than 45°), which reduces localization accuracy compared to the near-vertical top-down camera. Our full RGBs + tactile policy achieved up to 85$\%$ success in brushing, 70$\%$ in tong picking, and 90$\%$ in knife cutting, clearly outperforming all baselines (Tab.~\ref{tab:basline}). These results highlight why tactile sensing, combined with complementary visual perspectives, is essential for robust and contact-rich manipulation.

% ====================== TABLE ======================
\begin{table*}[t]
\centering
\scriptsize
\begin{tabularx}{\textwidth}{l *{15}{>{\centering\arraybackslash}X}}
\toprule
\textbf{} &
\multicolumn{5}{c}{\textbf{Brushing (num of demos)}} &
\multicolumn{5}{c}{\textbf{Tong picking (num of demos)}} &
\multicolumn{5}{c}{\textbf{Knife cutting (num of demos)}} \\
\cmidrule(lr){2-6}\cmidrule(lr){7-11}\cmidrule(lr){12-16}
& 20 & 40 & 60 & 80 & 100
& 20 & 40 & 60 & 80 & 100
& 20 & 40 & 60 & 80 & 100 \\
\midrule
Tactile Only
  & 0.00 & 0.00 & 0.00 & 0.05 & 0.10
  & 0.00 & 0.00 & 0.05 & 0.10 & 0.10
  & 0.00 & 0.00 & 0.10 & 0.15 & 0.20 \\
Egocentric RGB w/ Tactile
  & 0.00 & 0.00 & 0.00 & 0.20 & 0.40
  & 0.00 & 0.05 & 0.10 & 0.10 & 0.20
  & 0.05 & 0.15 & 0.10 & 0.10 & 0.15 \\
Top RGB w/ Tactile
  & 0.00 & 0.15 & 0.20 & 0.20 & 0.50
  & 0.05 & 0.30& 0.25& 0.30&0.35
  & \best{0.20} & \best{0.40} & \best{0.40} & 0.40 & 0.50 \\
RGBs only
  & 0.00 & 0.30 & 0.35 & 0.45 & 0.70
  & 0.10 & 0.10 & 0.30 & 0.40 & 0.45
  & 0.10 & 0.20 & 0.15 & 0.50 & 0.75 \\  
RGBs w/ Tactile (Ours)
  & \best{0.40} & \best{0.55} & \best{0.70} & \best{0.75} & \best{0.85}
  & \best{0.40} & \best{0.40} & \best{0.40} & \best{0.45} & \best{0.70}
  & 0.10 & \best{0.40} & \best{0.40} & \best{0.55} & \best{0.90} \\
\bottomrule
 \end{tabularx}
\caption{\textbf{Comparison of success rates across baselines.} 
We report single-seed results for all modality ablations in the table. To further assess robustness, we also run two additional seeds for RGB-only and our method and present their variance and statistical significance in Fig.~\ref{fig:t-test}. We bold the best results in each column.}
\label{tab:basline}
\vspace{-2em}
\end{table*}
Figure~\ref{fig:t-test} shows the success rate of policies trained with different proportions of demonstration data. Across all data regimes, policies trained with tactile sensing consistently match or outperform RGB-only baselines. The gap is most pronounced in low-data settings, where tactile input provides critical feedback for precise, contact-rich manipulation. We further run two additional seeds for RGB-only and our method, and the paired $t$-tests in Figure~\ref {fig:t-test} confirm that these improvements are statistically significant. These results highlight the value of tactile feedback not only for overall task success but also for reducing the number of demonstrations needed in imitation learning.

\begin{table}[!h]
    \vspace{-0.5em}
    \centering
    \scriptsize
    \setlength{\tabcolsep}{4pt}
    \renewcommand{\arraystretch}{0.9}
    \begin{tabular}{@{}lc lc@{}}
    \toprule
    \textbf{Method} & \textbf{Success Rate} & \textbf{Method} & \textbf{Success Rate} \\
    \midrule
    Egocentric RGB w/ Tactile & 0.10 & RGBs only              & 0.35 \\
    Top RGB w/ Tactile        & 0.20 & RGBs w/ Tactile (Ours) & 0.60 \\
    \bottomrule
    \end{tabular}
    \caption{Overall success rates for whole-task execution by chaining the best-performing policy checkpoint from each stage.}
    \label{tab:score-real}
    \vspace{-2.em}
\end{table}
We tested the high-level tool selector, which achieved 96.8$\%$ classification accuracy on the test set, demonstrating reliable selection of the appropriate tool for each subtask. We then evaluated the complete system by chaining the tool-specific policies to complete the full long-horizon task. As shown in Tab.~\ref{tab:score-real},  across 20 rollouts, the system achieved an average success rate of 60$\%$. Most failures were due to tofu slippage or encounters with out-of-distribution states not covered in the demonstrations, highlighting the need for broader training data and more systematic recovery strategies. Nevertheless, within the demonstration coverage, we qualitatively observe recovery from external perturbations: when the tofu is pushed away during execution, the visuotactile policy re-approaches it and continues the task, as shown in the supplementary video. Overall, these results show that tactile sensing plays a crucial role in scaling imitation learning to long-horizon daily activities. \Method\ successfully integrates tactile-enhanced policies to execute tasks such as cooking pan-fried tofu, where precision, robustness, and tool coordination are essential.

% \vspace{-0.5em}
\section{Related Work}
% \vspace{-0.5em}
\subsection{Tool Manipulation}
%Tool manipulation\cite{edsinger2007toward,zhang2008kinematic} has been an essential problem for understanding intelligence in cognitive science studies.
% To-do: cite some fairy tale([4], [3], [42], [19].)
To enable robots to understand and manipulate tools, prior work has explored task-oriented grasping and functional affordances through self-supervised learning \cite{fang2020learning,turpin2021gift}. However, these works mainly focus on how to utilize a single tool's different parts, i.e., how the tool should be grasped. \cite{lin2022planning}, \cite{shi2023robocook} further demonstrate long-horizon manipulation involving diverse tools through planning and learning.
In contrast, our work focuses on an integrated hardware and learning system that converts everyday household tools into interchangeable, tactile-instrumented end effectors, combines tool-matched teleoperation with tactile feedback for demonstration collection, and learns visuotactile policies for contact-rich manipulation.

Moreover, previous works \cite{fang2020learning, qin2020keto} try to learn the synergy between different tool grasping points and tasks using self-supervised learning methods. The learned latent representations could provide a structured and condensed understanding of tool objects. However, our method does not rely on tool-affordance representations; instead, it uses a mechanically simple custom tool changer to directly focus on the interaction and manipulation parts of the task.

\subsection{Tool Design for Robot Tool Usage}
% \vspace{-0.5em}
Several works aim to extend the range of tasks a single robot can perform by improving tool design and switching mechanisms. Industrial robots such as the UR5 employ dedicated tool changers that provide robust 
and precise switching performance. However, these systems are typically tailored to specific robots and tools, are expensive, and cannot easily accommodate actively articulated tools. 
Previous works have explored interchangeable surgical instruments \cite{mckinley2016interchangeable} and mechanically actuated tools compatible with parallel grippers \cite{hu2019designing}. In contrast, SwapBot provides a unified, low-cost system that supports both passive and articulated household tools, ensures secure coupling and reliable torque transmission for forceful, contact-rich manipulation, and integrates tactile sensing with tool-matched teleoperation for demonstration collection and downstream policy learning.
\vspace{-0.5em}
\subsection{Tactile Sensing for Manipulation}
Tactile sensors capture real-time force information and have been widely applied for dexterous, real-world manipulation. 
% They address the limitations of vision-based systems, such as occlusion and restricted fields of view. 
Visual-tactile systems, such as GelSight \cite{yuan2017gelsight}, use a soft reflective layer paired with an embedded camera to detect force magnitude and direction through subtle surface deformation. These systems provide high-resolution tactile feedback and have been demonstrated for a variety of robotic tasks, including fine-grained manipulation \cite{she2021cable}, material identification \cite{guo2023estimating}, and pose estimation \cite{higuera2024sparsh}. To extend tactile coverage over larger areas of a robot’s body, capacitive and resistive tactile skins are often used \cite{luo2024adaptive}. These skins typically use a three-layer stack of orthogonal electrodes with a dielectric/resistive layer, where electrode intersections act as pressure-sensing nodes. This scalable design supports seamless integration onto customized robotic components, such as grippers \cite{zlokapa2022integrated}, arms \cite{luo2021learning}, and full-body surfaces \cite{burns2022endowing}, enabling broader and more versatile tactile sensing in complex environments. Our system adopts scalable tactile sensor designs and customizes them to diverse tools via digital fabrication. Tactile data has also been integrated into policy training for a variety of manipulation tasks, often showing substantial performance gains. For example, sensors mounted on parallel grippers~\cite{huang20243d, zhao2025polytouch} provide valuable contact information that improves grasp stability and task success. However, such approaches are typically limited to specific gripper designs. In contrast, \Method\ enables more flexible measurement at the tool–object interface. 
%This provides a scalable way to instrument diverse tools with tactile feedback, supporting a wider range of manipulation capabilities.

% Second the tools in \cite{hu2019designing} don't connect tightly with the robot end-effector, causing the robot to fail at many manipulation tasks that require high torque to support the loads. In contrast, our hardware design ensures a tight connection between the robot end-effector and the tool adaptors, while remaining cost-effective.

%\subsection{Learning from Demonstration}
% The idea of learning from demonstration has a long history,

%In the data-driven (deep) imitation learning context, many of those works train one policy in an end-to-end manner to directly map observations to a sequence of actions or single step action~\cite{4813899,pmlr-v100-sieb20a,zhang2018deep}. 
%These methods do not show much generalization across novel objects and scenes. 

%\sandy{You don't really have a Discussion. What follows is a summary/limitations/future work => Conclusion.}
% \vspace{-0.5em}
\section{Conclusion}
We presented \Method, a modular, tool-centric manipulation system with integrated tactile sensing that improves demonstration collection and policy learning for long-horizon tasks. Tactile feedback proved crucial for both teleoperation quality and imitation learning robustness, especially with limited data. However, the current system has limitations: tactile sensors remain low resolution and prone to wear, sensor placement is fixed by human intuition, and policy learning depends on predefined task segmentation. Future work will explore more durable, higher-resolution sensing, data-driven strategies for sensor placement, and autonomous policy sequencing without manual subgoal definitions. We will also open-source the system to promote reproducibility and community adoption.
% \vspace{-0.8em}

%%%%%%%%%%%%%%%%%%%%%%%%%%%%%%%%%%%%%%%%%%%%%%%%%%%%%%%%%%%%%%%%%%%%%%%%%%%%%%%%

%%%%%%%%%%%%%%%%%%%%%%%%%%%%%%%%%%%%%%%%%%%%%%%%%%%%%%%%%%%%%%%%%%%%%%%%%%%%%%%%



\section*{ACKNOWLEDGMENT}
This work was (partially) funded by grants from the Office of Naval Research (\#N00014-24-S-B001 and STTR 25.A) and Amazon. We gratefully acknowledge gifts from Collaborative Robotics, Cruise, Grab, and others. We thank members of the Washington Embodied Intelligence and Robotics Development Lab, Wearable Intelligence Lab, and the Personal Robotics Lab for their thoughtful comments and feedback on versions of this draft.

%%%%%%%%%%%%%%%%%%%%%%%%%%%%%%%%%%%%%%%%%%%%%%%%%%%%%%%%%%%%%%%%%%%%%%%%%%%%%%%%

\bibliographystyle{ICRA/IEEEtran}
\bibliography{ICRA/IEEEabrv,bibliography}

\end{document}